\documentclass[twoside,11pt]{article}

\usepackage{amsmath,bm,paralist}
\usepackage{jmlr2e}

\def \v {\mathbf{v}}

\def \R {\mathbb{R}}

\def \N {\mathcal{N}}

\def \e {\mathbf{e}}

\def \Bh {\widehat{B}}

\def \Rt {\mathcal{R}}

\def \P {\mathcal{P}}

\def \uu {\mathbf{u}}
\def \E {\mathrm{E}}
\DeclareMathOperator*{\Pro}{P}

\DeclareMathOperator*{\argmax}{argmax}
\DeclareMathOperator*{\st}{s.t.}

\newtheorem{thm}{Theorem}

\newtheorem{cor}[thm]{Corollary}

\begin{document}

\title{Relative Error Bound Analysis for Nuclear Norm Regularized Matrix Completion}

\author{\name Lijun Zhang \email zhanglj@lamda.nju.edu.cn\\
       \addr National Key Laboratory for Novel Software Technology\\
       Nanjing University, Nanjing 210023, China
       \AND
       \name Tianbao Yang \email tianbao-yang@uiowa.edu\\
       \addr Department of Computer Science\\
       the University of Iowa, Iowa City, IA 52242, USA
       \AND
       \name Rong Jin \email rongjin@cse.msu.edu\\
       \addr Department of Computer Science and Engineering\\
        Michigan State University, East Lansing, MI 48824, USA
       \AND
       \name Zhi-Hua Zhou \email zhouzh@lamda.nju.edu.cn\\
       \addr National Key Laboratory for Novel Software Technology\\
       Nanjing University, Nanjing 210023, China}

\editor{}

\maketitle

\begin{abstract}
In this paper, we develop a \emph{relative} error bound for nuclear norm regularized matrix completion, with the focus on the completion of full-rank matrices. Under the assumption that the top eigenspaces of the target matrix are incoherent, we derive a relative upper bound for recovering the best low-rank approximation of the unknown matrix. Although multiple works have been devoted to analyzing the recovery error of full-rank matrix completion, their error bounds are usually additive, making it impossible to obtain the perfect recovery case and more generally difficult to leverage the skewed distribution of eigenvalues. Our analysis is built upon the optimality condition of the regularized formulation  and existing guarantees for low-rank matrix completion. To the best of our knowledge, this is the first relative bound that has been proved for the regularized formulation of matrix completion.
\end{abstract}


\section{Introduction}
Matrix completion is concerned with the problem of recovering an unknown matrix from a small fraction of its entries \citep{Candes:MC:2010}. Recently, the problem of low-rank matrix completion has received a great deal of interests due to the theoretical advances \citep{Candes:MC:2009,MC:Keshavan}, as well as its application to a wide range of real-world problems, including collaborative filtering \citep{Goldberg:1992:UCF}, sensor networks \citep{Biswas:2006:SPB}, computer vision \citep{NIPS2011_4419}, and machine learning \citep{ICML2011Jalali}.

Let $A$ be an unknown matrix of size $m\times n$, and without loss of generality, we assume $m \leq n$. The information available about $A$ is a sampled set of entries $A_{ij}$, $(i,j) \in \Omega$, where $\Omega$ is a subset of the complete set of entries $[m] \times [n]$. Our goal is to recover $A$ as precisely as possible. 
In a seminal work, \citet{Candes:MC:2009} assume that $A$ is low-rank, and propose to recover $A$ from the observed entries in $\Omega$ by solving the following nuclear norm minimization problem
\begin{equation} \label{eqn:first}
\begin{array}{ll}
\min & \|B\|_* \\
\st & B_{ij}=A_{ij} \ \forall (i,j) \in \Omega.
\end{array}
\end{equation}
Under the incoherence condition, they prove that with a high probability the solution to (\ref{eqn:first}) yields a perfect reconstruction of $A$,
provided that a sufficiently large number of entries are observed randomly.
When $A$ is of full rank, a similar nuclear norm minimization problem has been proposed. Suppose $A=Z+N$, where $Z$ is a low-rank matrix to recover, and $N$ is the residual matrix. \citet{MC:Noise:Candes} introduce the following problem for recovering $A$
\begin{equation} \label{eqn:noise}
\begin{array}{ll}
\min & \|B\|_* \\
\st & \sqrt{\sum_{(i,j) \in \Omega} (B_{ij}-A_{ij})^2} \leq \delta
\end{array}
\end{equation}
where $\delta$ is an upper bound for $\sqrt{\sum_{(i,j) \in \Omega} N_{ij}^2}$. Although a relative error bound has been established for (\ref{eqn:noise}) when $\delta$ is large enough \citep{MC:Noise:Candes}, the high computational cost with solving the optimization problem in (\ref{eqn:noise}), mostly due to the constraint and non-smooth objective function, makes it practically less attractive.

An alternative approach to (\ref{eqn:noise}) for matrix completion is to solve a nuclear norm regularized least squares problem
\begin{equation} \label{eqn:opt}
\min\limits_{B \in \R^{m\times n}} \quad \frac{1}{2} \sum_{(i,j) \in \Omega} (B_{ij}-A_{ij})^2 + \lambda\|B\|_{*}.
\end{equation}
This is the approach that is favored by practitioners because it can be solved significantly more efficiently than (\ref{eqn:noise}). In fact, a number of efficient optimization methods have been designed \citep{Ji:2009:AGM,AP:Nuclear,Trace:Norm:Reg,NIPS2012_4663,Nuclear:Active}. Using the accelerated gradient method \citep{Nesterov_Composite_new}, the convergence rate for solving (\ref{eqn:opt}) is $O(1/T^2)$, where $T$ is number of iterations, and can be even boosted to a linear convergence under mild conditions \citep{NIPS2013_4936}. In contrast, the convergence rate for (\ref{eqn:noise}) could be as low as $O(1/\sqrt{T})$.


Although (\ref{eqn:opt}) is computation-friendly, its recovery guarantee remains unclear. One may argue that (\ref{eqn:noise}) and (\ref{eqn:opt}) are equivalent by setting $\delta$ and $\lambda$ appropriately, but the exact correspondence between them is unknown in general. To bridge the gap between practice and theory, in this paper we provide a \emph{relative} error bound for the regularized formulation in (\ref{eqn:opt}). More specifically, assume $A$ is a matrix of full rank to be recovered. Let $A_r$ be the best rank-$r$ approximation of $A$, and $\widehat{A}$ be the matrix recovered from the observed entries in $\Omega$. A relative upper bound takes the following form
\begin{eqnarray}
\|\widehat{A} -A_r\|_F \leq U(r,m,n,|\Omega|) \|A-A_r\|_F \label{eqn:bound-1}
\end{eqnarray}
where $U(\cdot)$ is a function of $r$, $m$, $n$ and $|\Omega|$.~\footnote{By the triangle inequality $\|\widehat{A} -A\|_F \leq \|\widehat{A} -A_r\|_F + \|A -A_r\|_F$, a relative upper bound for recovering $A_r$ directly implies a relative upper bound for recovering $A$.} Note that this kind of bounds is very popular in compressive sensing \citep{CS:best:K} and low-rank matrix approximation \citep{Boutsidis:2009:IAA}.
Compared to the additive error bound, the key advantage of the relative error bound is that it bounds the error based on $\|A - A_r\|_F$, the approximation error between the original matrix $A$ and its low-rank approximation $A_r$. As a result, when $A$ is low-rank and $A - A_r = \mathbf{0}$, relative error bounds imply a perfect recovery of $A$, which will never be accomplished by additive bounds.


In this work, we are interested in bounding $\|B_* - A_r\|_F$ in the form of (\ref{eqn:bound-1}), where $B_*$ is the optimal solution to (\ref{eqn:opt}). Similar to previous studies, we assume that the top eigenspaces of $A$ satisfy the classical incoherence condition \citep{Candes:MC:2009}. Based on the celebrated result of low-rank matrix completion \citep{Recht:2011:SAM}, we derive an upper bound for $\|B_*-A_r\|_F$, which induces a relative upper bound under favored conditions. We summarize the key features of our results as follows:
\begin{compactitem}
      \item We present a general theorem that allows us to bound the recovery error of (\ref{eqn:opt}) for any $\lambda>0$. In contrast,  \citet{MC:Noise:Candes} only analyze the performance of (\ref{eqn:noise}) when $\delta \geq \sqrt{\sum_{(i,j) \in \Omega} N_{ij}^2}$.
   \item By choosing $\lambda$ appropriately, we obtain a relative upper bound of $O(\frac{mn \sqrt{r}}{|\Omega|}  \|A-A_r\|_F) $ in general, and a tighter bound of $O(\sqrt{\frac{mn r }{|\Omega|}} \|A-A_r\|_F )$ when  $A-A_r$ is flat, i.e., $\|A-A_r\|_\infty / \|A - A_r\|_F$ is not too large. Although \citet{MC:Koltchinskii} and \citet{RSC:Wainwright} have analyzed some variants of (\ref{eqn:opt}), their bounds are additive in the sense that they are not proportional to $ \|A-A_r\|_F$. To the best of our knowledge, this is the \emph{first} relative error bound for the nuclear norm regularized matrix completion.
   \item Our relative upper bound for (\ref{eqn:opt}) is tighter than that for (\ref{eqn:noise}) developed by \citet{MC:Noise:Candes}, and more general than those proved by  \citet{MC:Noise:Keshavan} and \citet{High-Rank:12} under different conditions.
   \item Compared to the additive upper bounds of other methods \citep{MC:Noise:Keshavan,MC:Koltchinskii,CB:MC:2011}, our relative upper bound is tighter when $\|A-A_r\|_F$ is small. In addition, our relative error bound implies the  perfect recovery case  when the target matrix $A$ is low-rank while the additive error never vanishes.
 \end{compactitem}

\textbf{Notations} For a matrix $X$, we use $\|X\|_*$, $\|X\|_F$, $\|X\|$, and $\|X\|_\infty$ to denote its nuclear norm, Frobenius norm, spectral norm, and the absolute value of the  largest element in magnitude, respectively,
\section{Related Work}
In this section, we provide a brief review of existing work.
\subsection{Low-rank Matrix Completion}
The mathematical study of matrix completion began with \citet{Candes:MC:2009}. Specifically, they have proved that if $A$ obeys the incoherence condition, $|\Omega| \geq C n^{6/5} r \log (n)$ is sufficient to ensure that with a high probability, $A$ is the unique solution to (\ref{eqn:first}), where $C$ is a constant independent from $r$, $m$, and $n$ \citep{Candes:MC:2009}. The lower bound for the size of $\Omega$ is subsequently improved to $n r \log^6(n)$ under a stronger assumption \citep{Candes:MC:2010}. These theoretical guarantees are without question great breakthroughs, but the proof techniques are highly involved. In two subsequent studies \citep{Recht:2011:SAM,TIT:MC:Gross}, the authors present a very elegant approach for analyzing (\ref{eqn:first}), and give slightly better bounds. For example, \citet{Recht:2011:SAM} improves the bound for $|\Omega|$ to $r n \log^2(n)$ and requires the weakest assumptions on $A$. The simplification of the analysis also leads to better understanding of matrix completion, and lays the foundations of the study in this paper.

In an alternative line of work,  \citet{MC:Keshavan} study matrix completion using a combination of spectral techniques and manifold optimization. The proposed algorithm named OPTSPACE, also achieves exact recovery if $|\Omega| \geq C n r \max(\log(n),r)$. However, the constant $C$ in their bound depends on many factors of $A$ such as the aspect ratio and the condition number.
After the pioneering work mentioned above, various algorithms and theories of matrix completion have been developed, including distributed matrix completion \citep{NIPS2011_DC}, matrix completion with side information \citep{NIPS2013_MC:Side}, 1-bit matrix completion \citep{JMLR:v14:cai13b}, noisy  matrix completion \citep{klopp2014}, coherent matrix completion \citep{Coherent:Matrix}, universal matrix completion \citep{Universal:Matrix}, and non-convex matrix completion \citep{Nonconvex:Matrix}, to name a few amongst many.
\subsection{Full-rank Matrix Completion} \label{sec:full:matrix}
Since existing studies for full-rank matrix completion differ significantly in their assumptions, their theoretical guarantees may not be directly comparable. In the following, we will state previous results in the most general form, and (if possible) characterize their behaviors with respect to $m$, $n$, $r$, and $|\Omega|$.

Denote the optimal solution of (\ref{eqn:noise}) by $\Bh$. Under the assumption $\delta \geq \sqrt{ \sum_{(i,j) \in \Omega} N_{ij}^2}$, Theorem 7 of \citet{MC:Noise:Candes} shows
\[
\|\Bh- Z\|_F\leq \left( 1+ m\sqrt{\frac{n}{|\Omega|}} \right) \delta.
\]
Let $Z=A_r$, $N=A-A_r$, and consider the optimal choice that $\delta =O\left(\sqrt{ \sum_{(i,j) \in \Omega} N_{ij}^2}\right)$. The above bound becomes
\begin{equation} \label{eqn:rel}
\|\Bh- A_r\|_F\leq \left( 1+ m\sqrt{\frac{n}{|\Omega|}} \right) \sqrt{ \sum_{(i,j) \in \Omega} (A-A_r)_{ij}^2}.
\end{equation}
One limitation of this work is that the theoretical guarantee is only valid when $\delta$ is sufficiently large. On the other hand, if we use a very large $\delta$, the upper bound becomes loose. Our result overcomes this limitation as our error bound holds for any positive regularization parameter $\lambda > 0$.

An investigation of OPTSPACE \citep{MC:Keshavan} for full-rank matrix completion is discussed in \citet{MC:Noise:Keshavan}. In particular, Theorem~1.1 of \citet{MC:Noise:Keshavan} implies the following additive upper bound
\begin{equation} \label{eqn:Keshavan}
O\left( \|A_r\|_{\infty}  m^{1/4} n^{5/4} \sqrt{\frac{r}{|\Omega|}} +   \frac{mn\sqrt{r}}{|\Omega|} \|U\| \right)
\end{equation}
where $U$ is some matrix that depends on $A-A_r$ and $\Omega$. Although it is possible to derive a relative upper bound from Theorem~1.2 of \citet{MC:Noise:Keshavan}, it requires very strong assumptions about the coherence, the aspect ratio ($n/m$), the condition number of $A_r$ and the $r$-th singular value of $A$. Thus, the bound derived from Theorem~1.2 of \citet{MC:Noise:Keshavan} is significantly more restricted than the bound proved here.

\citet{CB:MC:2011} study the problem of matrix completion from the view point of supervised learning. The optimization problem is formulated as least squares minimization subject to nuclear norm or max norm constraints. Their theoretical results follow from generic generalization guarantees based on the Rademacher complexity. Specifically, Theorem~6 of \citet{CB:MC:2011} implies the following additive upper bound
\begin{equation} \label{eqn:foygel}
O \left( \|A-A_r\|_F + n \sqrt{\frac{r m}{|\Omega|}} + \sqrt{ n \|A-A_r\|_F} \left(\frac{rm}{|\Omega|}\right)^{\frac{1}{4}} \right)
\end{equation}
where logarithmic factors are ignored. The derivation of (\ref{eqn:foygel}) is given in Appendix~\ref{sec:app}.

\citet{MC:Koltchinskii} have investigated a general trace regression model, which includes matrix completion as a special case. For matrix completion, they propose the following optimization problem
\[
\min\limits_{B \in \R^{m\times n}} \quad \frac{1}{2} \left\| B-  \frac{mn}{|\Omega|} \sum_{(i,j) \in \Omega} A_{ij} \e_{i} \e_{j}^\top \right\|_F^2  + \lambda\|B\|_{*}.
\]
Let $\widehat{B}$ be the optimal solution to the above problem. Under appropriate conditions, it has been proved that with a high probability \citep[Corollary 2]{MC:Koltchinskii}
\begin{equation} \label{eqn:skii}
\begin{split}
\|\widehat{B}-A\|_F^2 + \|\widehat{B}-X\|_F^2  \leq \|X-A\|_F^2 +  \frac{C m n^2 \log(n) \textrm{rank}(X)}{|\Omega|}
\end{split}
\end{equation}
for all $X \in \R^{m\times n}$. However, due to the presence of the second term in the upper bound, it is impossible to obtain a relative error bound.

\citet{RSC:Wainwright} have analyzed a variant of (\ref{eqn:opt}), which contains an additional $\ell_\infty$-norm constraint. Based on assumptions about the spikiness and rank of the target matrix, they derive the restricted strong convexity condition, and establish the following additive bound \citep[Theorem 2]{RSC:Wainwright}
\begin{equation} \label{eqn:right}
\begin{split}
& \max \left( \left(\frac{m n^2 \log n}{|\Omega|}\right)^{1/4}\sqrt{\left\|A-A_r\right\|_*} , n \sqrt{\frac{r m \log n}{|\Omega|}}  \right) \\
\leq & \max \left( \left(\frac{m^2 n^2 \log n}{|\Omega|}\right)^{1/4}\sqrt{\left\|A-A_r\right\|_F} , n \sqrt{\frac{r m \log n}{|\Omega|}}  \right).
\end{split}
\end{equation}
Thus, their optimization problem, assumptions and theoretical guarantees are all different from ours.

In a recent work, \citet{High-Rank:12} consider a high-rank matrix completion problem in which the columns of $A$ belong to a union of multiple low-rank subspaces. Under certain assumptions about the coherence as well as the geometrical arrangement of subspaces and the distribution of the columns in the subspaces, they develop a multi-step algorithm that is able to recover \emph{each} column of $A$ with a high probability, as long as $O(r n \log^2 (m))$ entries of $A$ are observed uniformly at random. However, the recovery guarantee of their algorithm for general full-rank matrices is unclear.

\section{Our Results}
We first describe  theoretical guarantees  and then provide some discussions.

\subsection{Theoretical Guarantees}
Let $U = [\uu_1, \ldots, \uu_r]$ and $V = [\v_1, \ldots, \v_r]$ be two matrices that contain the first $r$ left and right singular vectors of matrix $A$, respectively. Let $\e_i$ and $\e_j$ be the $i$-th and $j$-th standard basis in $\R^m$ and $\R^n$, respectively. Following the previous studies in matrix completion \citep{Candes:MC:2009,Recht:2011:SAM}, we define the coherence measure $\mu_0$ as
\[
\mu_0 = \max\left(\frac{m}{r}\max\limits_{1 \leq i \leq m} \|P_U\e_i\|^2, \frac{n}{r}\max\limits_{1 \leq j \leq n} \|P_V\e_j\|^2 \right)
\]
where $P_U = UU^{\top}$ and $P_V = VV^{\top}$ are two projection operators. We also define $\mu_1$ as
\[
\mu_1 = \max\limits_{i \in [m], j \in [n]} \sqrt{\frac{mn}{r}} \left|[UV^{\top}]_{ij} \right|.
\]
Define two projection operators $\P_T$ and $\P_{T^{\perp}}$ for matrices as
\[
\P_T(Z) = P_UZ + Z P_V - P_U Z P_V, \textrm{ and } \P_{T^{\perp}}(Z) =  (I - P_U)Z(I - P_V).
\]

We assume the indices are sampled uniformly with replacement, and thus $\Omega$ is a \emph{collection} that may contain duplicate indices. The linear operator $\Rt_{\Omega}: \R^{m\times n} \mapsto \R^{m\times n}$ is defined as
\[
\Rt_\Omega(Z) = \sum_{(i,j) \in \Omega} \langle \e_{i} \e_{j}^\top, Z \rangle \e_{i} \e_{j}^\top.
\]
To simplify the notation, we define
\[
\varepsilon = \|A - A_r\|_{F}.
\]

\subsubsection{A General Result}
Let $B_*$ be the optimal solution to (\ref{eqn:opt}). Based on the optimality condition of $B_*$ and the guarantee for low-rank matrix completion \citep{Recht:2011:SAM}, we obtain the following theorem.
\begin{thm} \label{thm:main}
Assume
\begin{equation} \label{eqn:omega:1}
|\Omega| \geq 114 \max(\mu_0, \mu_1^2) r (m+n) \beta \log^2(2n)
\end{equation}
for some $\beta>1$, and $n\geq 5$. With a probability at least $1-6\log(n)(m+n)^{2-2\beta}-n^{2-2\beta^{1/2}}$, we have
\[
\begin{split}
 \left\|\P_{T^{\perp}}(B_*)\right\|_F &\leq  \left\|\P_{T^{\perp}}(B_*)\right\|_* \leq  \frac{256 \beta \log^2 (n) \varepsilon^2}{9 \lambda} + \frac{3mn r \log(2n)  \lambda}{|\Omega|},\\
\|\P_T(A_r-B_*)\|_F  &\leq  \frac{16\log (n) \varepsilon}{3} \sqrt{\frac{2 \beta mn }{|\Omega|}} +  \frac{2 mn\lambda}{|\Omega|} \sqrt{3 r   \log(2n)} \\
&+64 \log (n) \sqrt{\frac{mn \beta}{6 |\Omega|}} \| \P_{T^\perp}(B_*)\|_F .
\end{split}
\]
\end{thm}
As can be seen, our upper bound is valid for any $\lambda>0$. In contrast, the upper bound for (\ref{eqn:noise}) in \citep{MC:Noise:Candes} is limited to the case $\delta \geq \sqrt{\langle \Rt_{\Omega}(A - A_r, A-A_r) \rangle}$.

By choosing $\lambda$ to minimize the upper bounds in the above theorem, we obtain the following corollary.
\begin{cor} \label{cor:relative} Under the condition in Theorem~\ref{thm:main}. Set
\begin{equation} \label{eqn:opt:lamda}
\lambda = \frac{16  \varepsilon}{3} \sqrt{\frac{ \beta  \log(2n) |\Omega| }{3 mn r  }}.
\end{equation}
With a probability at least $1-6\log(n)(m+n)^{2-2\beta}-n^{2-2\beta^{1/2}}$, we have
\[
\begin{split}
 \left\|\P_{T^{\perp}}(B_*)\right\|_F  &\leq   \frac{32 \log (2 n)}{3} \sqrt{ \frac{3 \beta mn r \log(2n)  }{|\Omega|}}  \varepsilon,\\
\|\P_T(A_r-B_*)\|_F & \leq  \left( 19 \log(2n) \sqrt{\frac{ \beta  mn  }{  |\Omega|}} +  \frac{2048 \beta \log^2 (2 n)mn }{3|\Omega|}   \sqrt{ \frac{ r \log(2n)  }{2 }}  \right) \varepsilon,
\end{split}
\]
and thus
\[
\left\|A_r-B_*\right\|_F  \leq O\left( \log (n) \sqrt{ \frac{ mn r \log(n)  }{|\Omega|}}  + \frac{mn \log^2(n)}{|\Omega|} \sqrt{r \log(n)}  \right) \varepsilon.
\]
\end{cor}
Corollary~\ref{cor:relative} shows that, with an appropriate choice of the parameter $\lambda$,  we can obtain a relative upper bound. One way to estimate $\lambda$ is to use the cross validation technique, an approach that is widely used in learning. More specifically, we can divide the observed entries into two separate sets: the training set and the validation set. We will use the training set to find the optimal solution to the recovered matrix, and use the validation set to determine the appropriate parameter $\lambda$. When (\ref{eqn:opt:lamda}) holds, we can also express the upper bounds in terms of $\lambda$. It is easy to verify that with a high probability, we have
\begin{equation} \label{eqn:upper:by:lamda}
\|A_r-B_*\|_F \leq O \left(  \frac{mnr \log^2(n)}{|\Omega| } \sqrt{\frac{mn}{|\Omega|}}  \right)\lambda.
\end{equation}
In the special case when $A$ is a rank-$r$ matrix, i.e., $A = A_r$, (\ref{eqn:upper:by:lamda}) implies the smaller the $\lambda$, the better the bound. In other words, we have $\|A-B_* \|_F \rightarrow 0$ as $\lambda \rightarrow 0$.

Finally, we note that whether the upper bound in Corollary~\ref{cor:relative} is tight remains open. Although both \citet[Theorem 6]{MC:Koltchinskii} and \citet[Theorem 3]{RSC:Wainwright} have established lower bounds for noisy full-rank matrix completion, their bounds become $0$ in the noisy-free setting. Thus, existing lower bounds cannot be used to examine the optimality of our result, and we will investigate the lower bound for noisy-free setting in the future.
\subsubsection{A Special Result with Tighter Bounds}
In the case that the residual matrix $A - A_r$ is not too spiky, in other words, $\|A-A_r\|_\infty / \|A - A_r\|_F$ is not too large, we obtain a tighter theorem as stated below.
\begin{thm} \label{thm:main:2}
Assume
\begin{equation} \label{eqn:omega:two}
|\Omega| \geq \max \left( 114 \max(\mu_0, \mu_1^2) r (m+n) \beta \log^2(2n), \frac{8 mn \|A-A_r\|_\infty^2}{3 \|A - A_r\|_F^2} \beta \log (n)\right)
\end{equation}
for some $\beta>1$, and $n\geq 5$. With a probability at least $1-6\log(n)(m+n)^{2-2\beta}-n^{2-2\beta^{1/2}}- n^{-\beta}$, we have
\[
\begin{split}
 \left\|\P_{T^{\perp}}(B_*)\right\|_F &\leq  \left\|\P_{T^{\perp}}(B_*)\right\|_* \leq     \frac{8|\Omega|\varepsilon^2}{mn \lambda} + \frac{3mn r \log(2n)  \lambda}{|\Omega|},\\
\|\P_T(A_r-B_*)\|_F  &\leq  4 \varepsilon +  \frac{2 mn\lambda}{|\Omega|} \sqrt{3 r   \log(2n)} +64 \log (n) \sqrt{\frac{mn \beta}{6 |\Omega|}} \| \P_{T^\perp}(B_*)\|_F .
\end{split}
\]
\end{thm}
In this theorem, we have two lower bounds for $|\Omega|$ in (\ref{eqn:omega:two}). If $A$ is low-rank, the second lower bound will vanish. Furthermore, it can be dropped when $(\|A-A_r\|^2_{\infty}/\|A - A_r\|_F^2) \leq O(r\log n/m)$, i.e., when the residual matrix does not concentrate on a small number of entries. Note that our flatness assumption is a condition over the residual matrix $A-A_r$. It is different from the $\ell_\infty$-norm constraint of \citet{RSC:Wainwright}, which is a requirement over the target matrix $A$.

By choosing $\lambda$ to minimize the upper bounds in Theorem~\ref{thm:main:2}, we obtain the following relative upper bounds.
\begin{cor} \label{cor:relative:two} Under the condition in Theorem~\ref{thm:main}. Set
\[
\lambda = \frac{2|\Omega| \varepsilon}{mn} \sqrt{\frac{2}{3r \log (2n)}}.
\]
With a probability at least $1-6\log(n)(m+n)^{2-2\beta}-n^{2-2\beta^{1/2}}- n^{-\beta}$, we have
\[
\begin{split}
 \left\|\P_{T^{\perp}}(B_*)\right\|_F  &\leq   4 \sqrt{6 r \log(2n)}\varepsilon,\\
\|\P_T(A_r-B_*)\|_F & \leq  \left(10+256  \sqrt{\frac{mn r \log^3(2n) \beta}{|\Omega|}} \right) \varepsilon,
\end{split}
\]
and thus
\[
\left\|A_r-B_*\right\|_F  \leq O\left( \sqrt{r \log (n)} + \sqrt{\frac{mn r  \log^3 (n)}{|\Omega|}} \right) \varepsilon.
\]
\end{cor}
As can be seen, the upper bound for $\|A_r-B_*\|_F $ in the above corollary is tighter than that in Corollary~\ref{cor:relative} by a factor of $\log(n)\sqrt{\frac{mn}{|\Omega|}}$.

\subsection{Comparisons}\label{Sec:com}
We compare our theoretical guarantees with previous results for matrix completion in this section. We focus on the practical scenario $|\Omega| \leq mn$, and for simplicity \emph{ignore} logarithmic factors.

The most comparable study is the relative upper bound derived by \citet{MC:Noise:Candes} for the constrained problem in (\ref{eqn:noise}), since their analysis also relies on the incoherence condition. In the general case, Corollary \ref{cor:relative} gives the following relative error bound
\[
\left\|A_r-B_*\right\|_F  \leq O\left(\frac{mn \sqrt{r}}{|\Omega|} \right) \varepsilon.
\]
From (\ref{eqn:rel}) in Section~\ref{sec:full:matrix}, we observe that \citet{MC:Noise:Candes} give the following bound
\[
\|A_r-\Bh\|_F\leq   O \left(  m \sqrt{\frac{n}{|\Omega|}}\right)\varepsilon.
\]
Because $|\Omega|\geq C nr$ for some constant $C$, we have
\[
m \sqrt{\frac{n}{|\Omega|}} \geq  \sqrt{C}\frac{mn \sqrt{ r}}{|\Omega|},
\]
which implies our bound is always tighter than that of \citet{MC:Noise:Candes}. In the case that (\ref{eqn:omega:two}) holds, Corollary~\ref{cor:relative:two}  indicates our relative error bound can be improved to
\[
\left\|A_r-B_*\right\|_F  \leq O\left(\sqrt{\frac{mn r }{|\Omega|}} \right) \varepsilon.
\]
Using Lemma~\ref{lem:2} in Section~\ref{sec:pro:thm2}, the error bound of \citet{MC:Noise:Candes} can also be improved and becomes
\[
\|A_r-\Bh\|_F\leq O \left(  m \sqrt{\frac{n}{|\Omega|}}\right) \sqrt{\langle \Rt_{\Omega}(A-A_r), A-A_r \rangle} \overset{\text{(\ref{eqn:lem:2})}}{\leq} O \left(  \sqrt{m} \varepsilon\right),
\]
which is again worse than our bound since $|\Omega|\geq C nr$.  Compared to the relative upper bounds of \citet{MC:Noise:Keshavan} and \citet{High-Rank:12}, our result is applicable to a more general case as their bounds only hold for a very restricted class of matrices.

Next, we compare our relative error bound with the additive bounds in previous studies \citep{MC:Noise:Keshavan,CB:MC:2011,MC:Koltchinskii,RSC:Wainwright}. Since those results are derived under different assumptions, the comparison should be treated conservatively. We remark that those assumptions are incomparable in general, since we can construct matrices to satisfy one assumption but violate others \citep[Setion 3.4.2]{RSC:Wainwright}. Our goal is to show that relative bounds could be tighter than additive bounds under certain conditions.

For brevity, we only provide the comparison using the tighter bound in Corollary~\ref{cor:relative:two}. Our relative bound $O(\sqrt{\frac{mn r }{|\Omega|}} \varepsilon )$ is tighter than the additive bound in (\ref{eqn:Keshavan}) derived by \citet{MC:Noise:Keshavan}, if $\varepsilon \leq O( n^{3/4}/m^{1/4 } )$ and also tighter than the additive bound in (\ref{eqn:foygel}) derived by \citet{CB:MC:2011}, if $\varepsilon \leq O(\sqrt{n})$. To compare with the additive bound in (\ref{eqn:skii}) derived by \citet{MC:Koltchinskii}, we set $X = A_r$ and have, with a high probability,
\[
\left\|\widehat{B}-A_r\right\|_F  \leq \varepsilon + O\left(\sqrt{\frac{mn^2 r }{|\Omega|}} \right)
\]
which is worse than our bound if $\varepsilon \leq O( \sqrt{n})$. Our bound is better than the additive bound in (\ref{eqn:right}) of \citet{RSC:Wainwright}, when
\[
\epsilon \leq \max \left( \frac{\sqrt{\Omega}}{r}, \sqrt{n} \right).
\]

Finally, we note that although our analysis is devoted to full-rank matrix completion, it can also be applied to noisy low-rank matrix completion. In this case, we have  $A=Z+N$, where $Z$ is a low-rank matrix and $N$ is the matrix of noise. As long as the eigenspaces of $Z$ satisfy the incoherence condition, our theoretical guarantees are valid by setting $A_r=Z$ and $\varepsilon=\|N\|_F$, even when $Z$ may not be the best rank-$r$ approximation of $A$.~\footnote{This can be easily verified because our analysis only requires the eigenspaces of $A_r$ satisfy the incoherence condition.} To compare with previous studies, let's assume $\|Z\|_F=1$ and entries of $N$ are independent sampled from $\N(0,\sigma^2)$ where $\sigma^2=1/(mn)$. Based on the concentration inequality for $\chi^2$-distributions \citep[Lemma 1]{Massart:2000}, with a high probability, $\varepsilon^2=O(\sigma^2 mn)=O(1)$. Then, our Corollary~\ref{cor:relative} and Corollary~\ref{cor:relative:two} imply that with a high probability
\[
\left\|Z-B_*\right\|_F  \leq O\left(  \frac{mn\sqrt{r}}{|\Omega|}   \right)  \textrm{ and }  \left\|Z-B_*\right\|_F  \leq O\left( \sqrt{\frac{mn r}{|\Omega|}} \right),
\]
respectively. In contrast, existing results for noisy low-rank matrix completion, e.g., Corollary 1 of \citet{RSC:Wainwright} and Theorem 7 of \citet{klopp2014}, have established an $O(\sqrt{rn /|\Omega|})$ bound. Thus, our bounds are loose for noisy low-rank matrix completion, which is probably because our analysis did not exploit the fact that entries of $N$ are i.i.d.~sampled.

\section{Analysis}
Although the current analysis is built upon the result from \citet{Recht:2011:SAM} that requires the incoherence assumption, it can be extended to support other assumptions for matrix completion. The key is to replace Theorem~\ref{thm:existing} below with the corresponding theorem derived under other assumptions. With appropriate replacement of Theorem~\ref{thm:existing}, we should still be able to obtain a relative error bound, of course with different dependence on $m$, $n$, $r$, and $|\Omega|$. For example, using the assumptions and theorems in \citet{Universal:Matrix}, we can generalize our result to a \emph{universal} guarantee for full-matrix completion. We leave the extension of our analysis to other assumptions as a future work.

\subsection{Sketch of the Proof} \label{eqn:sketch}
As we mentioned before, our analysis is built upon the existing theoretical guarantee for low-rank matrix completion, which is summarized below \citep{Recht:2011:SAM}.
\begin{thm} \label{thm:existing}
Suppose
\begin{equation} \label{eqn:omega:size}
|\Omega| \geq 32 \max(\mu_0, \mu_1^2) r (m+n) \beta \log^2(2n)
\end{equation}
for some $\beta > 1$. Then, with a probability at least $1-6\log(n)(m+n)^{2-2\beta}-n^{2-2\beta^{1/2}}$, the following statements are true:
\begin{compactitem}
\item
\begin{equation}\label{eqn:a-1}
\left\|\frac{mn}{|\Omega|} \P_T\Rt_{\Omega}\P_T - \P_T \right\|\leq \frac{1}{2}.
\end{equation}
\item
\begin{equation}\label{eqn:r:omega}
\|\Rt_\Omega\| \leq \frac{8}{3} \sqrt{\beta} \log (n).
\end{equation}
\item There exists a $Y \in \R^{m\times n}$ in the range of $\Rt_{\Omega}$ such that
\begin{align}
&\left\|\P_T(Y) - UV^{\top}\right\|_F  \leq \sqrt{\frac{r}{2n}}, \label{eqn:a-2} \\
& \left\|\P_{T^{\perp}(Y)}\right\| \leq \frac{1}{2} \label{eqn:a-3}, \\
&|\langle Y, A\rangle |  \leq  \sqrt{\frac{3mn r \log(2n)}{8|\Omega|}} \sqrt{ \langle \Rt_{\Omega}(A), A\rangle}, \label{eqn:a-4}
\end{align}
for all $ A \in \R^{m\times n}$.
\end{compactitem}
\end{thm}
The first part of above theorem contains concentration inequalities for the random linear operator $\P_T\Rt_{\Omega}\P_T$ and $\Rt_\Omega$, and the second part describes some important properties of a special matrix $Y$, which is used as an (approximate) dual certificate of (\ref{eqn:first}).

Next, we will examine the optimality of $B_*$ based on techniques from convex analysis, leading to the following theorem.
\begin{thm} \label{thm:opt}Let $B_*$ be the optimal solution to (\ref{eqn:opt}), we have
\begin{equation} \label{eqn:bound}
\lambda  \langle B_* - A_r,  UV^{\top}\rangle+ \lambda \|\P_{T^{\perp}}(B_*) \|_* \leq  \langle \Rt_{\Omega}(B_* - A), A_r-B_* \rangle.
\end{equation}
\end{thm}

Based on Theorems~\ref{thm:existing} and \ref{thm:opt}, we are ready to prove the main results. However, the analysis is a bit lengthy, so we split it into two parts, and will first show the following intermediate theorem.
\begin{thm} \label{thm:important}
Suppose (\ref{eqn:omega:size}) holds. With a probability at least $1-6\log(n)(m+n)^{2-2\beta}-n^{2-2\beta^{1/2}}$, we have
\begin{equation} \label{eqn:main:ineq}
\begin{split}
 & \frac{1}{2} \langle \Rt_{\Omega}( A_r-B_*), A_r-B_* \rangle + \frac{\lambda}{2}\left\|\P_{T^{\perp}}(B_*)\right\|_*  \\
 \leq & \lambda\sqrt{\frac{r}{2n}} \|\P_T(A_r-B_*)\|_{F} + \langle \Rt_{\Omega}(A-A_r), A-A_r \rangle + \frac{3mn r \log(2n) \lambda^2}{8|\Omega|}.
 \end{split}
\end{equation}
\end{thm}
Then, we can prove Theorem~\ref{thm:main} by further lower bounding and upper bounding the L.H.S.~and R.H.S.~of (\ref{eqn:main:ineq}), respectively.  If the residual matrix $A - A_r$ is not too spiky, we can apply Bernstein's inequality to derive a tighter bound for $\langle \Rt_{\Omega}(A-A_r), A-A_r \rangle$ and obtain Theorem~\ref{thm:main:2} in a similar way.

\subsection{Property of the Linear Operator $\Rt_\Omega$}
Before going to the detail, we first introduce a lemma that will be used throughout the analysis. Since $\Omega$ may contain duplicate indices, $\langle \Rt_\Omega(A), A \rangle \neq \| \Rt_\Omega(A)\|_F^2$ in general. We use the following lemma to take care of this issue.
\begin{lemma} \label{lem:suppot1}
\begin{align}
\langle \Rt_\Omega(A), A \rangle & \leq  \|\Rt_\Omega(A)\|_F^2,  \label{lem:pro:0}\\
|\langle \Rt_\Omega(A), B \rangle| & \leq  \sqrt{\langle \Rt_\Omega(A), A \rangle} \sqrt{\langle \Rt_\Omega(B), B\rangle} \label{lem:pro:1}
\end{align}
for all $A, B \in \R^{m \times n}$.
\end{lemma}
\begin{proof}
Denote the number of unique indices in $\Omega$ by $u$, and let $\Theta =\{(a_k,b_k)\}_{k=1}^u$ be a \emph{set} that contains all the unique indices in $\Omega$. Let $t_k$ denote the times that $(a_k,b_k)$ appears in $\Omega$. Then, we have
\[
 \langle \Rt_\Omega(A), A \rangle = \sum_{k=1}^u  t_k A^2_{a_k b_k} \leq \sum_{k=1}^u  t_k^2 A^2_{a_k b_k} = \|\Rt_\Omega(A)\|_F^2.
\]

To show (\ref{lem:pro:1}), we have
\[
 |\langle \Rt_\Omega(A), B \rangle| =  \left|\sum_{(i,j) \in \Omega} A_{ij} B_{ij} \right| \leq \sqrt{\sum_{(i,j) \in \Omega}  A_{ij}^2} \sqrt{\sum_{(i,j) \in \Omega}  B_{ij}^2}
=\sqrt{\langle \Rt_\Omega(A), A \rangle} \sqrt{\langle \Rt_\Omega(B), B\rangle}
\]
where the inequality is due to Cauchy--Schwarz inequality.
\end{proof}
\subsection{Proof of Theorem~\ref{thm:existing}}
Except for the last inequality in (\ref{eqn:a-4}), all the others can be found directly from Section 4 of \citet{Recht:2011:SAM}. Thus, we only provide the derivation of (\ref{eqn:a-4}), which is based on some intermediate results of \citet{Recht:2011:SAM}.

We first state those intermediate results. Following the construction of \citet[Section 4]{Recht:2011:SAM}, we partition $\Omega$ into $p$ partitions of size $q$. By assumption, we can choose
\[
 q \geq \frac{128}{3} \max(\mu_0,\mu_1^2) r (m+n) \beta \log(m+n)
\]
such that
\[
 p = \frac{\Omega}{q}=\frac{3}{4} \log 2n.
\]
Let $\Omega_j$ denote the set of indices corresponding to the $j$-th partition. We define $W_0 = U V^\top$,
\[
Y_k=\frac{mn}{q} \sum_{j=1}^k\Rt_{\Omega_j}(W_{j-1}), \textrm{ and }  W_k=U V^\top-\P_T(Y_k)
\]
for $k=1,\ldots,p$. Then, we set $Y=Y_p$. \citet{Recht:2011:SAM} has proved that with a probability at least $1-6\log(n)(m+n)^{2-2\beta}-n^{2-2\beta^{1/2}}$,
\begin{align}
& \|W_k\|_F  \leq 2^{-k} \sqrt{r}, \label{eqn:a-5} \\
& \left\| \frac{mn}{q} \P_T \Rt_{\Omega_k} \P_T - \P_T\right\|\leq\frac{1}{2}, \label{eqn:a-6}
\end{align}
for $k=1,\ldots,p$.

We  proceed to prove (\ref{eqn:a-4}). Since $W_j = \P_T(W_j)$, we have
\begin{equation}\label{eqn:a-7}
\begin{split}
& \frac{mn}{q} \langle \Rt_{\Omega_j}(W_j), W_j \rangle=  \left\langle W_j, \frac{mn}{q} \P_T\Rt_{\Omega_j}\P_T(W_j) \right \rangle \\
\overset{\text{(\ref{eqn:a-6})}}{\leq}  & \frac{3}{2} \left\|\P_T(W_j)\right\|_F^2 = \frac{3}{2} \|W_j\|_F^2 \overset{\text{(\ref{eqn:a-5})}}{\leq} \frac{3r}{2} 4^{-j}.
\end{split}
\end{equation}
Then,
\[
\begin{split}
|\langle Y, A\rangle|\leq & \frac{mn}{q} \sum_{j=1}^p \left|\langle \Rt_{\Omega_j}(W_{j-1}), A \rangle \right|  \\
\overset{\text{(\ref{lem:pro:1})}}{\leq} & \frac{mn}{q} \sum_{j=1}^p \sqrt{\langle \Rt_{\Omega_j}(W_{j-1}), W_{j-1}\rangle} \sqrt{\langle \Rt_{\Omega_j}(A), A\rangle} \\
\leq & \frac{mn}{q} \sqrt{\sum_{j=1}^p \langle \Rt_{\Omega_j}(W_{j-1}), W_{j-1}\rangle} \sqrt{\sum_{j=1}^p \langle \Rt_{\Omega_j}(A), A\rangle} \\
= & \sqrt{\frac{mn}{q}} \sqrt{\sum_{j=1}^p \left \langle \frac{mn}{q} \Rt_{\Omega_j}(W_{j-1}), W_{j-1} \right\rangle} \sqrt{ \langle \Rt_{\Omega}(A), A\rangle } \\
\overset{\text{(\ref{eqn:a-7})}}{\leq} & \sqrt{\frac{mn}{q}} \sqrt{ \langle \Rt_{\Omega}(A), A\rangle } \sqrt{\frac{3r}{2} \sum_{j=1}^p 4^{-j} }\\
\leq & \sqrt{\frac{mnr}{2q}} \sqrt{ \langle \Rt_{\Omega}(A), A\rangle } =\sqrt{\frac{mnpr}{2|\Omega|}} \sqrt{ \langle \Rt_{\Omega}(A), A\rangle }. \\
\end{split}
\]

\subsection{Proof of Theorem~\ref{thm:opt}}
Since $B_*$ is the optimal solution to (\ref{eqn:opt}), we have
\begin{equation} \label{eqn:opt:cond}
\langle  \Rt_{\Omega}(B_* - A) + \lambda E,  A_r - B_* \rangle \geq 0
\end{equation}
where $E \in  \partial \|B_*\|_*$ is \emph{certain} subgradient of $\|\cdot\|_*$ evaluated at $B_*$. Let $F \in  \partial \|A_r\|_*$ be \emph{any} subgradient of $\|\cdot\|_*$ evaluated at $A_r$. From the property of convexity, we have
\begin{equation} \label{eqn:conv:prop}
\langle B_*- A_r, E-F \rangle \geq 0.
\end{equation}
From (\ref{eqn:opt:cond}) and (\ref{eqn:conv:prop}), we get
\begin{equation} \label{eqn:1}
\langle \Rt_{\Omega}(B_* - A)+ \lambda F,   A_r - B_* \rangle  \geq  0.
\end{equation}
Next, we consider bounding $\lambda \langle F,  A_r - B_* \rangle$. From previous studies \citep{Candes:MC:2009}, we know that the set of subgradients of $\|A_r\|_{*}$ takes the following form:
\[
\partial \|A_r\|_*=\left\{UV^{\top}+W:  W \in \R^{m \times n}, U^\top W=0, WV=0, \|W\| \leq 1\right\}.
\]
Thus, we can choose
\[
F=UV^{\top}+ \P_{T^{\perp}}(N),
\]
where $N=\argmax_{\|X\|\leq 1} \langle \P_{T^{\perp}}(B_*), X \rangle$.
Then, it is easy to verify that
\begin{equation} \label{eqn:2}
\begin{split}
 \langle B_* - A_r, F \rangle  =& \langle B_* - A_r, UV^{\top} \rangle + \langle B_* - A_r, \P_{T^{\perp}}(N) \rangle \\
= & \langle B_* - A_r,  UV^{\top}\rangle+ \|\P_{T^{\perp}}(B_*) \|_* .
\end{split}
\end{equation}
We complete the proof by combining (\ref{eqn:1}) and (\ref{eqn:2}).

\subsection{Proof of Theorem~\ref{thm:important}} \label{sec:pro:important}
We continue the proof by lower bounding $\langle B_* - A_r,  UV^{\top}\rangle$ in (\ref{eqn:bound}) of Theorem~\ref{thm:opt}. To this end, we need the matrix $Y$ given in Theorem~\ref{thm:existing}.
\[
\begin{split}
\langle B_* - A_r, UV^{\top}\rangle  = & \langle B_* - A_r, UV^{\top} - Y\rangle + \langle B_* - A_r, Y\rangle \\
 = & \langle  B_* - A_r, UV^{\top} - \P_T(Y) \rangle + \langle A_r-B_* , \P_{T^{\perp}}(Y)\rangle+ \langle  B_* - A_r, Y\rangle.
 \end{split}
\]
Next, we bound the last three terms by utilizing the conclusions in Theorem~\ref{thm:existing}.
\[
\begin{split}
 &\langle  B_* - A_r, UV^{\top} - \P_T(Y) \rangle = \langle \P_T(B_* - A_r), UV^{\top} - \P_T(Y)\rangle \\
\geq & -\|\P_T(B_* - A_r)\|_F\|UV^{\top} - \P_T(Y)\|_F  \overset{\text{(\ref{eqn:a-2})}}{\geq}  -\sqrt{\frac{r}{2n}} \|\P_T( A_r-B_*)\|_{F}.
\end{split}
\]
\[
\begin{split}
&\langle  A_r-B_*, \P_{T^{\perp}}(Y)\rangle = \langle \P_{T^{\perp}}( A_r-B_*), \P_{T^{\perp}}(Y)\rangle\\
=&\langle \P_{T^{\perp}}(-B_*), \P_{T^{\perp}}(Y)\rangle \geq  -\|\P_{T^{\perp}}(B_*)\|_* \|\P_{T^{\perp}}(Y)\| \overset{\text{(\ref{eqn:a-3})}}{\geq}  - \frac{1}{2}\left\|\P_{T^{\perp}}(B_*)\right\|_*.
\end{split}
\]
\[
\begin{split}
 \langle  B_* - A_r, Y\rangle  \overset{\text{(\ref{eqn:a-4})}}{\geq}  - \sqrt{\frac{3mn r \log(2n)}{8|\Omega|}} \sqrt{ \langle \Rt_{\Omega}(A_r-B_*), A_r-B_*\rangle}.
\end{split}
\]
Putting the above inequalities together, we have
\begin{equation} \label{eqn:3}
\begin{split}
\langle B_* - A_r, UV^{\top}\rangle \geq & -\sqrt{\frac{r}{2n}} \|\P_T( A_r-B_*)\|_{F} - \frac{1}{2}\left\|\P_{T^{\perp}}(B_*)\right\|_*\\
&- \sqrt{\frac{3mn r \log(2n)}{8|\Omega|}} \sqrt{ \langle \Rt_{\Omega}(A_r-B_*), A_r-B_*\rangle }. \\
 \end{split}
\end{equation}

Substituting (\ref{eqn:3}) into (\ref{eqn:bound}) and rearranging, we get
\begin{equation} \label{eqn:4}
\begin{split}
 & \langle \Rt_{\Omega}( A_r-B_*), A_r-B_* \rangle + \frac{\lambda}{2}\left\|\P_{T^{\perp}}(B_*)\right\|_*  \\
 \leq & \langle \Rt_{\Omega}(A_r-A), A_r-B_* \rangle + \lambda\sqrt{\frac{r}{2n}} \|\P_T(A_r-B_*)\|_{F}   \\
 &+ \lambda \sqrt{\frac{3mn r \log(2n)}{8|\Omega|}} \sqrt{ \langle \Rt_{\Omega}(A_r-B_*), A_r-B_*\rangle}\\
 \overset{\text{(\ref{lem:pro:1})}}{\leq} & \sqrt{\langle \Rt_{\Omega}(A-A_r), A-A_r \rangle}  \sqrt{\langle \Rt_{\Omega}(A_r-B_*), A_r-B_* \rangle}+ \lambda\sqrt{\frac{r}{2n}} \|\P_T(A_r-B_*)\|_{F}\\
 &+ \lambda \sqrt{\frac{3mn r \log(2n)}{8|\Omega|}} \sqrt{ \langle \Rt_{\Omega}(A_r-B_*), A_r-B_*\rangle}.
\end{split}
\end{equation}

From the the basic inequality $\frac{1}{4} \alpha^2 - \alpha  \beta +  \beta^2 \geq 0$, we have
\begin{equation} \label{eqn:5}
\begin{split}
&\sqrt{\langle \Rt_{\Omega}(A-A_r), A-A_r \rangle}  \sqrt{\langle \Rt_{\Omega}(A_r-B_*), A_r-B_* \rangle} \\
\leq &  \frac{1}{4} \langle \Rt_{\Omega}( A_r-B_*), A_r-B_* \rangle +  \langle \Rt_{\Omega}(A-A_r), A-A_r \rangle,
\end{split}
\end{equation}
\begin{equation} \label{eqn:6}
\begin{split}
&\lambda \sqrt{\frac{3mn r \log(2n)}{8|\Omega|}} \sqrt{ \langle \Rt_{\Omega}(A_r-B_*), A_r-B_*\rangle} \\
\leq & \frac{1}{4} \langle \Rt_{\Omega}( A_r-B_*), A_r-B_* \rangle +  \lambda^2 \frac{3mn r \log(2n)}{8|\Omega|}.
\end{split}
\end{equation}
We complete the proof by summing (\ref{eqn:4}), (\ref{eqn:5}), and (\ref{eqn:6}) together.
\subsection{Proof of Theorem~\ref{thm:main}}
The lower bound of $|\Omega|$ in (\ref{eqn:omega:1}) is due to Theorem~\ref{thm:existing}, but we use a larger constant ($114$ instead of $32$) to ensure
\begin{equation}\label{eqn:explain:omega}
 \frac{8 \log (n)}{3} \sqrt{\frac{rm\beta}{|\Omega|}}   \leq \frac{1}{4}
\end{equation}
which is used later.

Based on Lemma~\ref{lem:suppot1} and Theorem~\ref{thm:existing}, we have
\[
\sqrt{\langle \Rt_{\Omega}(A - A_r), A-A_r \rangle} \overset{\text{(\ref{lem:pro:0})}}{\leq }  \|\Rt_\Omega(A-A_r)\|_F \overset{\text{(\ref{eqn:r:omega})}}{\leq }\frac{8}{3} \sqrt{\beta}\log (n) \varepsilon.
\]
Substituting the above inequality into (\ref{eqn:main:ineq}), we have
\begin{equation} \label{eqn:main:ineq:diff}
\frac{1}{2} \langle \Rt_{\Omega}( A_r-B_*), A_r-B_* \rangle + \frac{\lambda}{2}\left\|\P_{T^{\perp}}(B_*)\right\|_*   \leq  \lambda\sqrt{\frac{r}{2n}} \|\P_T(A_r-B_*)\|_{F} + \Gamma,
\end{equation}
where
\begin{equation} \label{eqn:gamma}
\Gamma=\frac{64 \beta \log^2 (n) \varepsilon^2}{9}  + \frac{3mn r \log(2n) \lambda^2}{8|\Omega|}.
\end{equation}

\subsubsection{Upper Bound for $\|\P_{T^{\perp}}(B_*)\|_F$} We upper bound $\|\P_T(A_r-B_*)\|_{F}^2$ in (\ref{eqn:main:ineq:diff}) by
\[
\|\P_T(A_r-B_*)\|_{F}^2 = \langle \P_T(A_r-B_*),  A_r-B_* \rangle \overset{\text{(\ref{eqn:a-1})}}{\leq} 2 \frac{mn}{|\Omega|} \langle \P_T\Rt_{\Omega}\P_T(A_r-B_*), A_r-B_*\rangle.
\]
Plugging the above inequality in (\ref{eqn:main:ineq:diff}), we have
\begin{equation} \label{eqn:main:1}
\begin{split}
 & \frac{1}{2} \langle \Rt_{\Omega}( A_r-B_*), A_r-B_* \rangle + \frac{\lambda}{2}\left\|\P_{T^{\perp}}(B_*)\right\|_*  \\
 \leq  & \lambda\sqrt{\frac{rm}{|\Omega|}} \sqrt{\langle \P_T\Rt_{\Omega}\P_T(A_r-B_*), A_r-B_*\rangle}  + \Gamma.
\end{split}
\end{equation}
Since $\P_T+ \P_{T^\perp}=I$, we have
\begin{equation} \label{eqn:main:2}
\begin{split}
 &\frac{1}{2} \langle \Rt_{\Omega}( A_r-B_*), A_r-B_* \rangle\\
 =& \frac{1}{2} \underbrace{\langle \P_T\Rt_{\Omega}\P_T(A_r-B_*), A_r-B_*  \rangle}_{:=\Theta^2} + \frac{1}{2} \underbrace{\langle \P_{T^\perp}\Rt_{\Omega}\P_{T^\perp}( B_*), B_*  \rangle}_{:=\Lambda^2}\\
& - \langle  \Rt_{\Omega}( \P_T(A_r-B_*)), \P_{T^\perp}(B_*) \rangle \\
\overset{\text{(\ref{lem:pro:1})}}{\geq} & \frac{1}{2} \Theta^2  + \frac{1}{2} \Lambda^2   - \Theta \Lambda = \frac{1}{2} (\Theta-\Lambda)^2.
\end{split}
\end{equation}
Substituting (\ref{eqn:main:2}) into (\ref{eqn:main:1}), we have
\[
 \frac{1}{2} (\Theta-\Lambda)^2 + \frac{\lambda}{2}\left\|\P_{T^{\perp}}(B_*)\right\|_* \leq \lambda\sqrt{\frac{rm}{|\Omega|}} \Theta + \Gamma.
\]
Combining with the fact
\[
 \frac{1}{2} (\Theta-\Lambda)^2 -\lambda\sqrt{\frac{rm}{|\Omega|}} \Theta + \lambda\sqrt{\frac{rm}{|\Omega|}} \Lambda + \frac{rm \lambda^2}{2|\Omega|} = \frac{1}{2} \left(\Theta-\Lambda-  \lambda\sqrt{\frac{rm}{|\Omega|}}\right)^2  \geq 0
\]
we have
\begin{equation} \label{eqn:old:1}
\begin{split}
\frac{\lambda}{2}\left\|\P_{T^{\perp}}(B_*)\right\|_*  \leq & \lambda\sqrt{\frac{rm}{|\Omega|}} \sqrt{\langle \P_{T^\perp}\Rt_{\Omega} \P_{T^\perp}(B_*), B_*\rangle}+ \frac{ rm \lambda^2}{2|\Omega|} + \Gamma \\
 \overset{\text{(\ref{lem:pro:0})}}{\leq} & \lambda\sqrt{\frac{rm}{|\Omega|}} \|\Rt_{\Omega} \P_{T^\perp}(B_*)\|_F+ \frac{ rm \lambda^2}{2|\Omega|} + \Gamma\\
  \overset{\text{(\ref{eqn:r:omega})}}{\leq} &  \frac{8\lambda \log (n)}{3} \sqrt{\frac{rm\beta}{|\Omega|}}  \| \P_{T^\perp}(B_*)\|_F+ \frac{ rm \lambda^2}{2|\Omega|} + \Gamma\\
\overset{\text{(\ref{eqn:explain:omega})}}{\leq} & \frac{\lambda}{4} \| \P_{T^\perp}(B_*)\|_F+ \frac{ rm \lambda^2}{2|\Omega|} + \Gamma \\
\overset{\text{(\ref{eqn:gamma})}}{\leq} & \frac{\lambda}{4} \| \P_{T^\perp}(B_*)\|_*+ \frac{64 \beta \log^2 (n) \varepsilon^2}{9}+ \frac{3mn r \log(2n)  \lambda^2}{4|\Omega|}  \\
\end{split}
\end{equation}
where in the last line we use the fact
\[
 \frac{ 1 }{2}  \leq \frac{3n  \log(2n) }{8} , \ \forall n \geq 2.
\]

From (\ref{eqn:old:1}), we immediately have
\[
 \left\|\P_{T^{\perp}}(B_*)\right\|_* \leq  \frac{256 \beta \log^2 (n) \varepsilon^2}{9 \lambda} + \frac{3mn r \log(2n)  \lambda}{|\Omega|}.
\]

\subsubsection{Upper Bound for $\|\P_T(A_r-B_*)\|_F$}
 Similar to (\ref{eqn:main:2}), we have
\[
\begin{split}
& \frac{1}{2} \langle \Rt_{\Omega}( A_r-B_*), A_r-B_* \rangle \\
\overset{\text{(\ref{lem:pro:1})}}{\geq} & \frac{1}{2} \langle \P_T\Rt_{\Omega}\P_T(A_r-B_*), A_r-B_*  \rangle + \frac{1}{2} \Lambda^2  - \sqrt{\langle \P_T \Rt_{\Omega} \P_T(A_r-B_*), A_r-B_*  \rangle} \Lambda \\
\overset{\text{(\ref{eqn:a-1})}}{\geq} & \frac{|\Omega|}{4 mn} \|  \P_T(A_r-B_*)\|_F^2+ \frac{1}{2}  \Lambda^2 - \sqrt{\frac{3|\Omega|}{2mn} }\|  \P_T(A_r-B_*)\|_F  \Lambda\\
\end{split}
\]
where
\[
\Lambda=\sqrt{\langle \P_{T^\perp}\Rt_{\Omega}\P_{T^\perp}( B_*), B_*  \rangle} \overset{\text{(\ref{lem:pro:0})}}{\leq}  \| \Rt_{\Omega}\P_{T^\perp}( B_*)\|_F \overset{\text{(\ref{eqn:r:omega})}}{\leq} \frac{8}{3} \sqrt{\beta} \log (n)  \| \P_{T^\perp}( B_*)\|_F.
\]

By plugging the above inequalities into (\ref{eqn:main:ineq:diff}), we have
\[
\begin{split}
& \frac{|\Omega|}{4 mn} \|  \P_T(A_r-B_*)\|_F^2+ \frac{1}{2}  \Lambda^2 + \frac{\lambda}{2}\left\|\P_{T^{\perp}}(B_*)\right\|_*  \\
\leq & \lambda\sqrt{\frac{r}{2n}} \|  \P_T(A_r-B_*)\|_F +\Gamma+8 \log (n) \sqrt{\frac{|\Omega| \beta}{6mn}} \| \P_{T^\perp}(B_*)\|_F     \|  \P_T(A_r-B_*)\|_F \\
\end{split}
\]
and thus
\begin{equation} \label{eqn:old:2}
\begin{split}
 \|  \P_T(A_r-B_*)\|_F^2 \overset{\text{(\ref{eqn:gamma})}}{\leq}  &    \frac{2 m\lambda \sqrt{2rn}}{|\Omega|} \| \P_T(A_r-B_*)\|_F +\frac{256 \beta mn \log^2 (n) \varepsilon^2}{9|\Omega|}
 + \frac{3 r  m^2n^2\lambda^2 \log(2n)}{2|\Omega|^2} \\
& +32 \log (n) \sqrt{\frac{mn \beta}{6 |\Omega|}} \| \P_{T^\perp}(B_*)\|_F  \|  \P_T(A_r-B_*)\|_F.
\end{split}
\end{equation}

Recall that
\[
x^2 \leq bx +c \Rightarrow x \leq 2b + \sqrt{2c}.
\]
From (\ref{eqn:old:2}), we have
\[
\begin{split}
\|  \P_T(A_r-B_*)\|_F \leq &  \frac{4 m\lambda \sqrt{2rn}}{|\Omega|}  +64 \log (n) \sqrt{\frac{mn \beta}{6 |\Omega|}} \| \P_{T^\perp}(B_*)\|_F \\
 & +   \frac{16\log (n) \varepsilon}{3} \sqrt{\frac{2 \beta mn }{|\Omega|}} +  \frac{mn\lambda}{|\Omega|} \sqrt{3 r   \log(2n)} .
 \end{split}
\]
We complete the proof by noticing
\[
4 \sqrt{2n} \leq n \sqrt{3\log(2n)}, \ \forall n \geq 5.
\]

\subsection{Proof of Theorem~\ref{thm:main:2}} \label{sec:pro:thm2}
With the second lower bound of $|\Omega|$ in  (\ref{eqn:omega:two}), we can prove the following upper bound for $\langle \Rt_{\Omega}(A - A_r), A-A_r \rangle$.
\begin{lemma} \label{lem:2}
Suppose
\begin{equation} \label{eqn:omega:2}
 |\Omega| \geq      \frac{8 mn \|A-A_r\|_\infty^2}{3 \|A - A_r\|_F^2} \beta \log (n)
\end{equation}
for some $\beta > 1$.  Then, with a probability at least $1 - n^{-\beta}$, we have
\begin{equation} \label{eqn:lem:2}
\sqrt{\langle \Rt_{\Omega}(A - A_r), A-A_r \rangle} \leq \varepsilon \sqrt{\frac{2|\Omega|}{mn}}.
\end{equation}
\end{lemma}
\begin{proof}
For each index $(a_k,b_k) \in \Omega$, we define a random variable
\[
\xi_k= \langle \e_{a_k} \e_{b_k}^\top, A-A_r\rangle^2 - \frac{1}{mn}\|A - A_r\|_F^2.
\]
Then, it is easy to verify that
\[
\begin{split}
\E[\xi_k]  =& 0, \\
|\xi_k| = & \left|\langle \e_{a_k} \e_{b_k}^\top, A-A_r\rangle^2 - \frac{1}{mn}\|A - A_r\|_F^2 \right| \\
\leq &  \max\left(\langle\e_{a_k} \e_{b_k}^\top, A-A_r\rangle^2, \frac{1}{mn}\|A - A_r\|_F^2 \right) \leq  \|A-A_r\|_\infty^2,\\
\E [\xi_k^2] =& \E \left[\langle \e_{a_k} \e_{b_k}^\top, A-A_r\rangle^4 \right]- \frac{1}{m^2n^2}\|A - A_r\|_F^4 \\
\leq & \frac{1}{m n} \sum_{i,j} [A - A_r]_{ij}^4  \leq  \frac{1}{m n} \|A-A_r\|_\infty^2 \|A-A_r\|_F^2.
\end{split}
\]

From Bernstein's inequality, we have
\[
\begin{split}
& \Pro \left[ \langle \Rt_{\Omega}(A - A_r), A-A_r \rangle \geq 2 \frac{|\Omega|}{mn}\|A - A_r\|_F^2 \right] \\
=&  \Pro \left[ \sum_{k=1}^{|\Omega|} \xi_k \geq \frac{|\Omega|}{mn}\|A - A_r\|_F^2 \right] \leq \exp\left( - \frac{3 |\Omega|}{8 \|A-A_r\|_\infty^2} \frac{1}{mn}\|A - A_r\|_F^2 \right) \overset{\text{(\ref{eqn:omega:2})}}{\leq}   n^{-\beta}.
\end{split}
\]
\end{proof}

Following the derivation of (\ref{eqn:main:ineq:diff}), we have
\[
\frac{1}{2} \langle \Rt_{\Omega}( A_r-B_*), A_r-B_* \rangle + \frac{\lambda}{2}\left\|\P_{T^{\perp}}(B_*)\right\|_*   \leq  \lambda\sqrt{\frac{r}{2n}} \|\P_T(A_r-B_*)\|_{F} + \Gamma',
\]
where
\[
\Gamma'=\frac{2|\Omega|\varepsilon^2}{mn} + \frac{3mn r \log(2n)  \lambda^2}{8|\Omega|}.
\]
The rest of the analysis is almost identical to that of Theorem~\ref{thm:main}. In particular, (\ref{eqn:old:1}) becomes
\[
\frac{\lambda}{2}\left\|\P_{T^{\perp}}(B_*)\right\|_*  \leq  \frac{\lambda}{4} \| \P_{T^\perp}(B_*)\|_*+ \frac{2|\Omega|\varepsilon^2}{mn} + \frac{3mn r \log(2n)  \lambda^2}{4|\Omega|},
\]
and (\ref{eqn:old:2}) becomes
\[
\begin{split}
 \|  \P_T(A_r-B_*)\|_F^2 \leq  &    \frac{2 m\lambda \sqrt{2rn}}{|\Omega|} \| \P_T(A_r-B_*)\|_F + 8\varepsilon^2 + \frac{3 r  m^2n^2\lambda^2 \log(2n)}{2|\Omega|^2} \\
& +32 \log (n) \sqrt{\frac{mn \beta}{6 |\Omega|}} \| \P_{T^\perp}(B_*)\|_F     \|  \P_T(A_r-B_*)\|_F.
\end{split}
\]

 A complete proof can be found in an early version of this paper \citep{MC:Full:Early}.
\section{Conclusion and Future Work}
In this paper, we develop a relative error bound for the nuclear norm regularized matrix completion, under the assumption that the top eigenspaces of the target matrix are incoherent. To the best of our knowledge, this is the first work toward relative error bound for nuclear norm regularized matrix completion, and an extensive comparison shows that our bound is tighter than previous results under favored conditions.

In many real-world applications, it is appropriate to assume the observed entries are corrupted by noise. As we discussed in the end of Section~\ref{Sec:com}, it is  possible to extend our analysis to the noisy case. More specifically, let $N$ be the matrix of noise. We just need to add $\langle \Rt_{\Omega}(N), A_r-B_* \rangle$  to the R.H.S.~of (\ref{eqn:bound}), which leads to an additional term $\sqrt{\langle \Rt_\Omega(N), N \rangle} \sqrt{\langle \Rt_\Omega(A_r-B_*), A_r-B_*\rangle}$ in the R.H.S.~of (\ref{eqn:4}). The rest of the proof is almost the same, and finally we will obtain an upper bound that depends on both $A-A_r$ and  $N$.

\appendix
\section{Derivation of (\ref{eqn:foygel})} \label{sec:app}
Let $\widehat{B}$ be the solution found by the algorithm in \citet{CB:MC:2011}. Let $\epsilon > 0$ be the mean-squared reconstruction error. Using the notations in this paper, (5) of \citet{CB:MC:2011} becomes
\begin{equation}\label{eqn:prove:1}
\frac{1}{mn} \|\widehat{B}-A\|_F^2 \leq \frac{1}{mn} \|A-A_r\|_F^2 + \epsilon
\end{equation}
under the condition
\begin{equation}\label{eqn:cond:1}
|\Omega| \geq O\left( \frac{r(n+m)}{\epsilon^2} \left(\epsilon+ \frac{1}{mn} \|A-A_r\|_F^2\right) \right)
\end{equation}
where logarithmic factors are ignored. (\ref{eqn:cond:1}) can be rewritten as
\[
|\Omega|  \epsilon^2 \geq O\left(r(n+m) \epsilon +  \frac{r(n+m)}{mn} \|A-A_r\|_F^2 \right).
\]
Since we assume $m \leq n$, it can be further simplified to
\[
|\Omega|  \epsilon^2 \geq O\left(rn \epsilon +  \frac{r}{m} \|A-A_r\|_F^2 \right).
\]

Let consider the optimal case, i.e.,
\[
|\Omega|  \epsilon^2 = C rn \epsilon +  C \frac{r}{m} \|A-A_r\|_F^2
\]
for some constant $C>0$. Then, we have
\[
\begin{split}
& \epsilon=\frac{C rn + \sqrt{C^2 r^2n^2 + 4|\Omega|  C \frac{r}{m} \|A-A_r\|_F^2  } }{2 |\Omega| } \\
 \leq &\frac{C rn + \sqrt{|\Omega|  C \frac{r}{m} \|A-A_r\|_F^2  } }{|\Omega| } =O\left( \frac{rn}{|\Omega|} + \sqrt{  \frac{r}{m |\Omega|} \|A-A_r\|_F^2  } \right).
\end{split}
\]
As a result, (\ref{eqn:prove:1}) becomes
\[
\frac{1}{mn} \|\widehat{B}-A\|_F^2 \leq O\left( \frac{1}{mn} \|A-A_r\|_F^2 + \frac{rn}{|\Omega|} + \sqrt{  \frac{r}{m |\Omega|} \|A-A_r\|_F^2}  \right)
\]
which implies
\[
 \|\widehat{B}-A\|_F^2 \leq O\left(\|A-A_r\|_F^2 + \frac{r m n^2}{|\Omega|} + \sqrt{  \frac{r m n^2}{ |\Omega|} \|A-A_r\|_F^2}  \right)
\]
from which we obtain (\ref{eqn:foygel}).

\bibliography{ref}

\vskip 0.2in

\end{document}